\documentclass[letterpaper, 10 pt, conference]{ieeeconf}  % Comment this line out if you need a4paper

\IEEEoverridecommandlockouts                              % This command is only needed if 
\usepackage{graphicx}
\usepackage{amsmath}
\usepackage{amssymb}
\usepackage{booktabs}
\usepackage{float}
\usepackage{cuted}
\usepackage{capt-of}
\usepackage{xcolor}
\usepackage{enumerate}
\usepackage{tabularx}
\usepackage{subcaption}
\usepackage{pifont}
\usepackage{multirow}

\def\etal{\textit{et al}.}

\def\method{LDM-ISP}

\title{\LARGE \bf
LDM-ISP: Enhancing Neural ISP for Low Light with Latent Diffusion Models
}

\author{Qiang Wen, Zhefan Rao, Yazhou Xing$^{*}$ and Qifeng Chen$^{*}$% <-this % stops a space
\thanks{$*$ denotes joint corresponding authors.}
\thanks{Qiang Wen (qwenab@connect.ust.hk), Yazhou Xing (yxingag@connect.ust.hk), Zhefan Rao (zraoac@connect.ust.hk) and  Qifeng Chen (cqf@ust.hk) are with the Department of Computer Science and Engineering, The Hong Kong University of Science and Technology, Hong Kong, China.}
}

\begin{document}

\maketitle
\thispagestyle{empty}
\pagestyle{empty}

%%%%%%%%%%%%%%%%%%%%%%%%%%%%%%%%%%%%%%%%%%%%%%%%%%%%%%%%%%%%%%%%%%%%%%%%%%%%%%%%
\begin{abstract}

  Enhancing a low-light noisy \textbf{RAW} image into a well-exposed and clean \textbf{sRGB} image is a significant challenge for modern digital cameras. Prior approaches have difficulties in recovering fine-grained details and true colors of the scene under extremely low-light environments due to near-to-zero SNR. Meanwhile,  diffusion models have shown significant progress towards general domain image generation. In this paper, we propose to leverage the pre-trained latent diffusion model to perform the neural ISP for enhancing extremely low-light images. Specifically, to tailor the pre-trained latent diffusion model to operate on the RAW domain, we train a set of lightweight taming modules to inject the RAW information into the diffusion denoising process via modulating the intermediate features of UNet. We further observe different roles of UNet denoising and decoder reconstruction in the latent diffusion model, which inspires us to decompose the low-light image enhancement task into latent-space low-frequency content generation and decoding-phase high-frequency detail maintenance. Through extensive experiments on representative datasets, we demonstrate our simple design not only achieves state-of-the-art performance in quantitative evaluations but also shows significant superiority in visual comparisons over strong baselines, which highlight the effectiveness of powerful generative priors for neural ISP under extremely low-light environments.

\end{abstract}

%%%%%%%%%%%%%%%%%%%%%%%%%%%%%%%%%%%%%%%%%%%%%%%%%%%%%%%%%%%%%%%%%%%%%%%%%%%%%%%%
\section{INTRODUCTION}

\begin{figure}[h]
\centering
\includegraphics[width=0.9\linewidth]{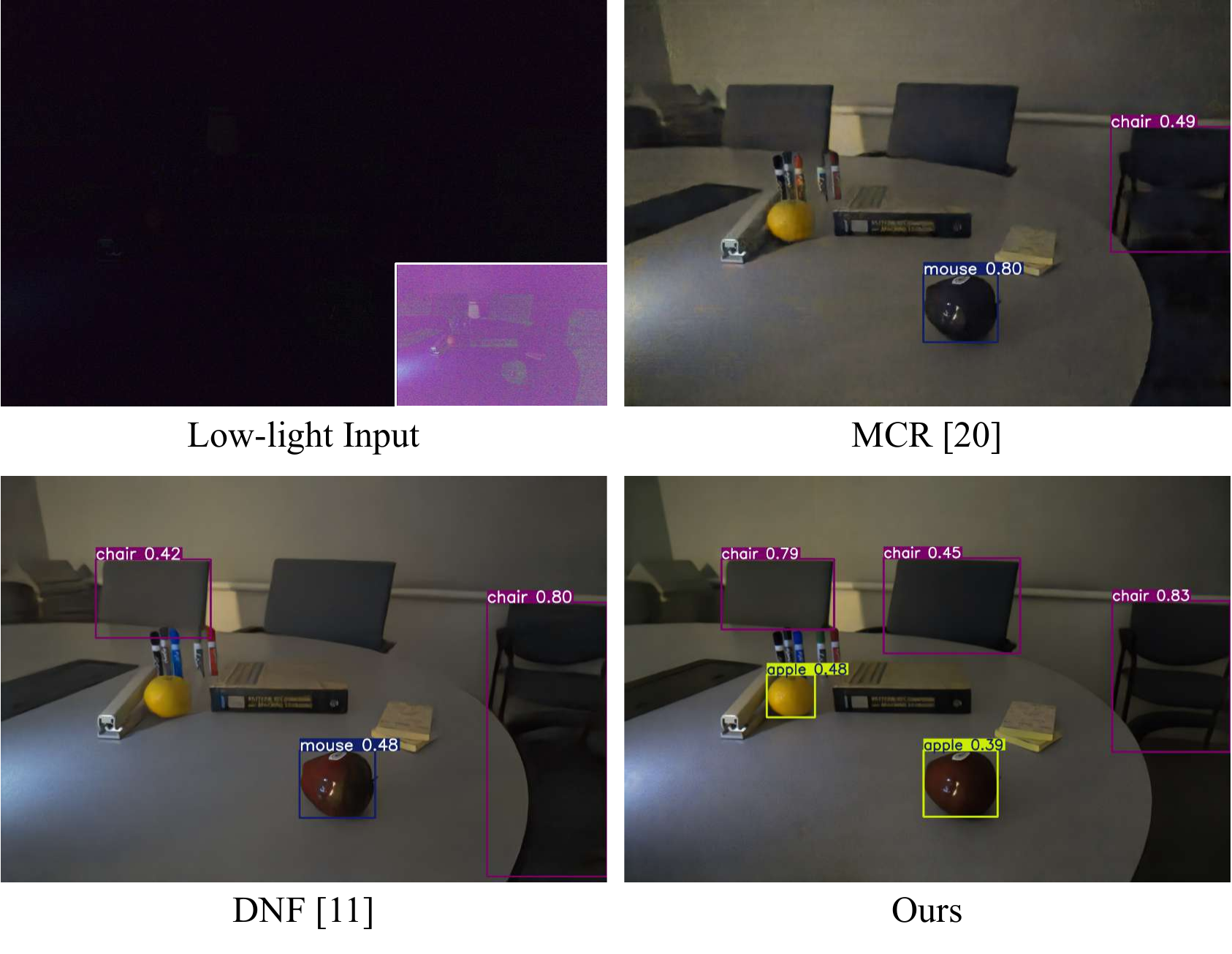}
\caption{The detection results obtained from YOLOv8~\cite{yolov8} demonstrate significant improvements from the proposed method. The proposed method yields clearer low-light enhancements, facilitating more accurate detection results.}
\label{fig: detection_visual}
\end{figure}

As the demand for robotic systems and surveillance technologies in low-light environments continues to rise, the importance of effective low-light image enhancement is becoming increasingly critical. Poor lighting conditions often lead to significant challenges in object detection, including increased noise and reduced visibility, which can severely impair performance. Enhancing low-light images is crucial for improving detection accuracy and reliability, enabling these systems to function effectively in diverse settings as shown in Fig.~\ref{fig: detection_visual}.
% Enhancing low-light photos to well-exposed and clean RGB images not only requires reducing a significant amount of noise but also grapples with challenges related to detail and structure recovery, color correction, and careful joint optimization of many other operations. Thus, imaging in extremely low-light environments has been a long-standing challenging problem for modern digital cameras.
 %
Standard approaches such as increasing the ISO, extending exposure time, and enlarging the aperture during capturing, all have inevitable limitations and thus cannot solve the problem in practice.
Learning-based methods, on the other hand, have been making surprising progress toward effective low-light imaging systems~\cite{chen2018learning, jin2023lighting, zhang2023towards, wei2021physics, wang2023exposurediffusion, yu2018deepexposure, chen2019seeing, jiang2019learning, zhu2020eemefn, huang2022towards, jincvpr23dnf}. However, most of the existing approaches learn the low-light enhancement model from relatively limited paired data and the enhanced RGB images often suffer from detail deficiency and color deviation issues.
 %
% In this paper, we leverage the powerful generative priors from a pre-trained latent diffusion model~\cite{rombach2022high} to enhance the neural ISP for low-light imaging.

Due to the significant practical value of low-light imaging, there have been 
many approaches to solve the task, which can be categorized into RGB-based enhancement~\cite{li2021learning, li2021low, lv2018mbllen, zamir2020learning, Chen2018Retinex, yang2020fidelity, xu2020learning} and RAW-based enhancement~\cite{chen2018learning, chen2019seeing, wei2021physics, jincvpr23dnf, wang2023exposurediffusion, jin2023lighting, zhang2023towards}.
RGB data, however, contains much less signal than its unprocessed RAW counterparts, and thus the performance of RGB-based enhancement is limited, especially under extremely low-light conditions.
To alleviate this issue, Chen~\etal~\cite{chen2018learning} propose to learn the entire ISP pipeline with paired low-light RAW and well-exposed sRGB data.
Dong~\etal~\cite{dong2022abandoning} present to simulate the monochrome raw image from the colored raw image to compensate the photon number of the dark raw images. 
DNF~\cite{jincvpr23dnf} improves the neural ISP through decoupling the noise-to-clean and RAW-to-sRGB process, and utilizing the feature-level information of noise-to-clean as feedback to the RAW-to-sRGB process. 
Although the learned RAW-based low-light enhancement has been made effective progress, existing approaches still learn the enhancement process on limited pairs of RAW-sRGB data, and thus the resultant models can hardly perform well in extremely low-light cases. 

% \begin{figure*}[!t]
% \includegraphics[width=\textwidth]{../figs/Teaser_ICRA.pdf}
% \caption{An extremely low-light real-world image from SID Dataset~\cite{chen2018learning} with a ×300 amplification ratio. The image processed by our proposed method exhibits clear improvement in structural content, showcasing finer details compared to previous works.($\dagger$ denotes the RAW-to-sRGB method)}
% \end{figure*}

In this work, we propose the LDM-ISP, which leverages the powerful generative priors of the pre-trained latent diffusion model (i.e., Stable Diffusion) to enhance the neural ISP for low light.
We use one end-to-end network to simulate the entire ISP pipeline, which learns denoising, color transformations, demosaicing, and detail enhancement all in one network. 
Designing such an end-to-end ISP is not a trivial problem. First, Stable Diffusion is designed for content generation in the RGB domain and it is questionable how to tailor the pre-trained Stable Diffusion to the RAW domain.
A naive way is to fine-tune the entire diffusion model parameters, which could, however, lead to catastrophic forgetting of the abundant priors~\cite{luo2023empirical, gao2023ddgr}.
Second, the low-light image enhancement task requires different levels of content enhancements, such as structural recovery, detail enhancement, and color correction.
It is also challenging to seamlessly merge the ISP enhancement to the different stages of the diffusion denoising process. 

To overcome these challenges, instead of directly finetuning the parameters of the latent diffusion model, \method~aims at inserting a set of trainable taming modules, whose key part is the spatial feature transformation (SFT) layer~\cite{wang2018recovering} to modulate the intermediate features of the latent diffusion model. 
Meanwhile, we observe that there are dedicated generative priors in different portions of the latent diffusion model, and thus we propose to employ 2D discrete wavelet transforms to divide the LLIE task into two sub-tasks: low-frequency content generation and high-frequency detail maintenance.
To be specific, \method~employs 2D discrete wavelet transforms on the low-light noisy RAW image to extract its low-frequency and high-frequency sub-bands.
Then, the low-frequency sub-band is fed into taming modules to steer the structural content generation while the high-frequency is adopted to maintain the precise and clear details in the final sRGB output.
Feeding low- and high-frequency subbands separately allows us to skillfully exploit dedicated generate priors in different portions of the latent diffusion model.
By training the proposed taming modules on a small-scale LLIE dataset, \method~finds an optimized balance between image fidelity and perceptual quality, which allows it to achieve state-of-the-art LLIE performance by generating high-quality and artifact-free images.

Our contributions are summarized as follows.
\begin{itemize}
 \renewcommand{\labelitemi}{\textbullet}
\item We are the first to propose an enhanced neural ISP for low light with the powerful generative priors of the latent diffusion model.

\item Instead of training or finetuning a diffusion model on a specific LLIE dataset, the proposed method \method~only needs to insert a set of trainable taming modules into a pre-trained latent diffusion model to generate high-quality outputs. This approach effectively tailors the latent diffusion model to the RAW domain.

\item To maximally exploit dedicated generative priors in different portions of the latent diffusion model, we further propose to divide the LLIE task into low-frequency content generation and high-frequency detail maintenance through the 2D discrete wavelet transforms (DWT) on the input RAW image.

\item Extensive evaluations conducted on three widely-used real-world datasets show the proposed method not only achieves state-of-the-art performance according to quantitative metrics but also significantly outperforms previous works in perceptual quality.
\end{itemize}

\section{Related Work}

\subsection{Learning-based Image Signal Processing}
Deep learning has shown its advantages in learning the entire ISP process~\cite{schwartz2018deepisp, liang2021cameranet, zamir2020cycleisp, zhang2021learning, xing2021invertible}. DeepISP~\cite{schwartz2018deepisp} proposes one of the first CNN-based ISPs to replace the traditional human-tuned ISP operations, including denoising, demosaicking, color correction, white balance, and photo finishing steps. Instead of utilizing one neural network to simulate the entire ISP pipeline, CameraNet~\cite{liang2021cameranet} presents a two-stage model for better RGB rendering performance. CycleISP~\cite{zamir2020cycleisp} learns the bidirectional mapping of RGB-to-RAW and RAW-to-RGB to recover the RAW images from sRGB input. Recently, Xing~\etal~\cite{xing2021invertible} proposes to utilize one single invertible neural network to learn an invertible ISP that can do both the RGB rendering and RAW reconstruction. However, most of the existing learning-based ISPs are designed for general photo-capturing conditions, and their performance under extremely low-light environments is unsatisfying. In this work, we aim to enhance the neural ISP for low light, with the abundant generative priors of pre-trained Stable Diffusion.

\subsection{RAW-based Low-light Image Enhancement}
For enhancing low-light RAW images, numerous studies have showcased the promising performance of deep learning-based methods.
These previous studies can be divided into two categories: RAW-to-RAW methods~\cite{wei2021physics, zhang2023towards, wang2023exposurediffusion, jin2023lighting} and RAW-to-sRGB methods~\cite{chen2018learning, chen2019seeing, huang2022towards, dong2022abandoning, zhu2020eemefn, lamba2021restoring, maharjan2019improving}.
RAW-to-RAW works focus on proposing denoising strategies on the RAW domain.
With carefully designed noise modeling strategies~\cite{wei2021physics, zhang2023towards} or denoising methods~\cite{wang2023exposurediffusion, jin2023lighting}, such works provide denoised RAW images and apply a known ISP on them for the final visualization.
RAW-to-sRGB works focus on tackling denoising and ISP tasks for low-light images.
Recently, such works have shown significantly superior LLIE performance by designing multi-stage methods~\cite{jincvpr23dnf, dong2022abandoning} that tackle noise-to-clean and RAW-to-sRGB tasks in separate stages, compared to single-stage method~\cite{chen2018learning, chen2019seeing, maharjan2019improving}.
This further indicates that there is a domain gap between denoising and ISP tasks, which makes the RAW-to-sRGB task more challenging than the RAW-to-RAW.

Compared to RAW-to-RAW methods, RAW-to-sRGB methods show a distinct advantage when the camera ISP is unknown or confidential.
Chen~\etal~\cite{chen2018learning} propose the first RAW-to-sRGB UNet-based method SID with a RAW dataset, which contains 5096 RAW data pairs for tackling the LLIE problem.
Zhu~\etal~\cite{zhu2020eemefn} propose a multi-stage method by fusing multi-exposure images to avoid the color bias issue.
Xu~\etal~\cite{xu2020learning} propose a multi-stage method based on a frequency-based decomposition to improve the detail enhancement.
However, both methods suffer from error accumulation across stages~\cite{jincvpr23dnf}.
Dong~\etal~\cite{dong2022abandoning} propose a multi-stage method MCR that first predicts a monochrome raw image for information restoration and then performs LLIE.
To address the challenge of the domain ambiguity issue in simultaneous denoising and ISP, Jin~\etal~\cite{jincvpr23dnf} propose a method DNF to tackle denoising and ISP tasks in separate stages.
Although these multi-stage methods show promising LLIE performance, they still grapple with color correction and denoising in extremely challenging scenarios.
The main reason for this shortcoming is that current training datasets~\cite{chen2018learning, wei2021physics, zhang2023towards, dong2022abandoning} contain limited scenes, which impedes the generalization ability of these learning-based methods.

\section{Approach}
\begin{figure*}[!t]
\centering
\includegraphics[width=\linewidth]{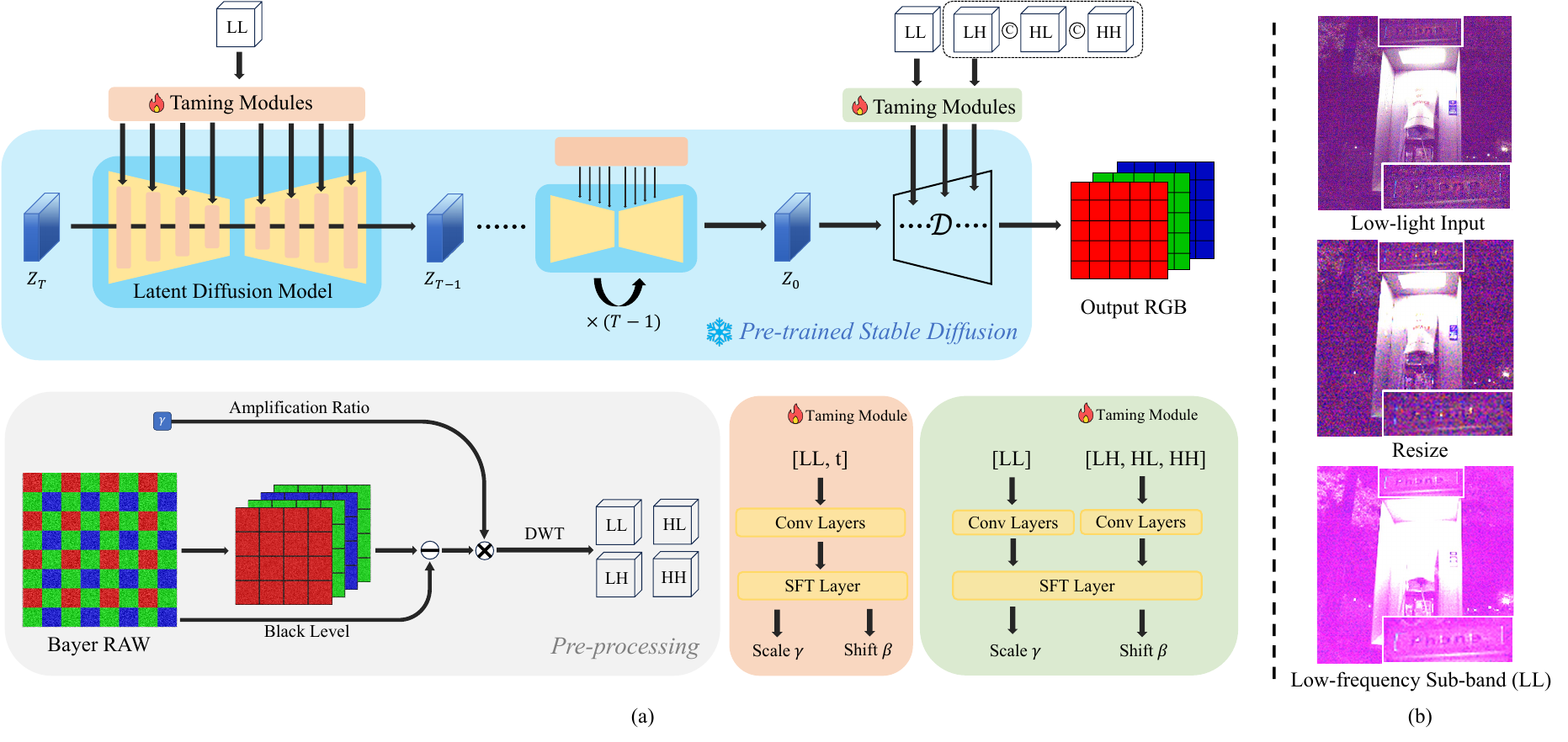}
\caption{(a) The overview of our proposed LLIE method, \textbf{\method}. The Bayer RAW image is processed by similar operations mentioned in~\cite{chen2018learning}. A series of 2D discrete wavelet transforms (DWT) is applied to the processed image for capturing the low-frequency (LL) and high-frequency subbands (LH, HL, HH). The low-frequency subband (LL) serves to modulate the feature at each layer in the UNet. Specifically, each feature has its corresponding taming module, whose key part is an SFT layer, to map the sub-band into a pair of scale $\gamma$ and shift $\beta$ parameters. Similar to the low-frequency taming, the features of the decoder $\mathcal{D}$ are modulated by another set of taming modules, where the LL sub-band is mapped to the scale $\gamma$ and the concatenation of LH, HL, HH is mapped to the shift $\beta$. All parameters from the pre-trained Stable Diffusion are \textbf{frozen} and only taming modules are \textbf{trainable}. (b) The low-frequency sub-band, extracted using 2D discrete wavelet transforms (DWT), reveals clearer structural information compared to the low-light input.}
\label{fig: framework}
\end{figure*}

An overview of the proposed method is shown in Fig.~\ref{fig: framework} (a).
Given a low-light noisy RAW image, we first process it to match the illumination to our target sRGB image.
% Given a low-light noisy RAW image, we first process it to match the illumination to our target sRGB image: $X = (\text{RAW} - \text{Black Level}) \times \gamma$, where $\gamma$ represents the user-defined amplification ratio.
 %
Then, we aim at taming the latent diffusion model to perform the image signal processing pipeline (ISP) with restoration for over-dark areas and denoising for the processed image.

\begin{figure}[!h]
\centering
\includegraphics[width=\linewidth]{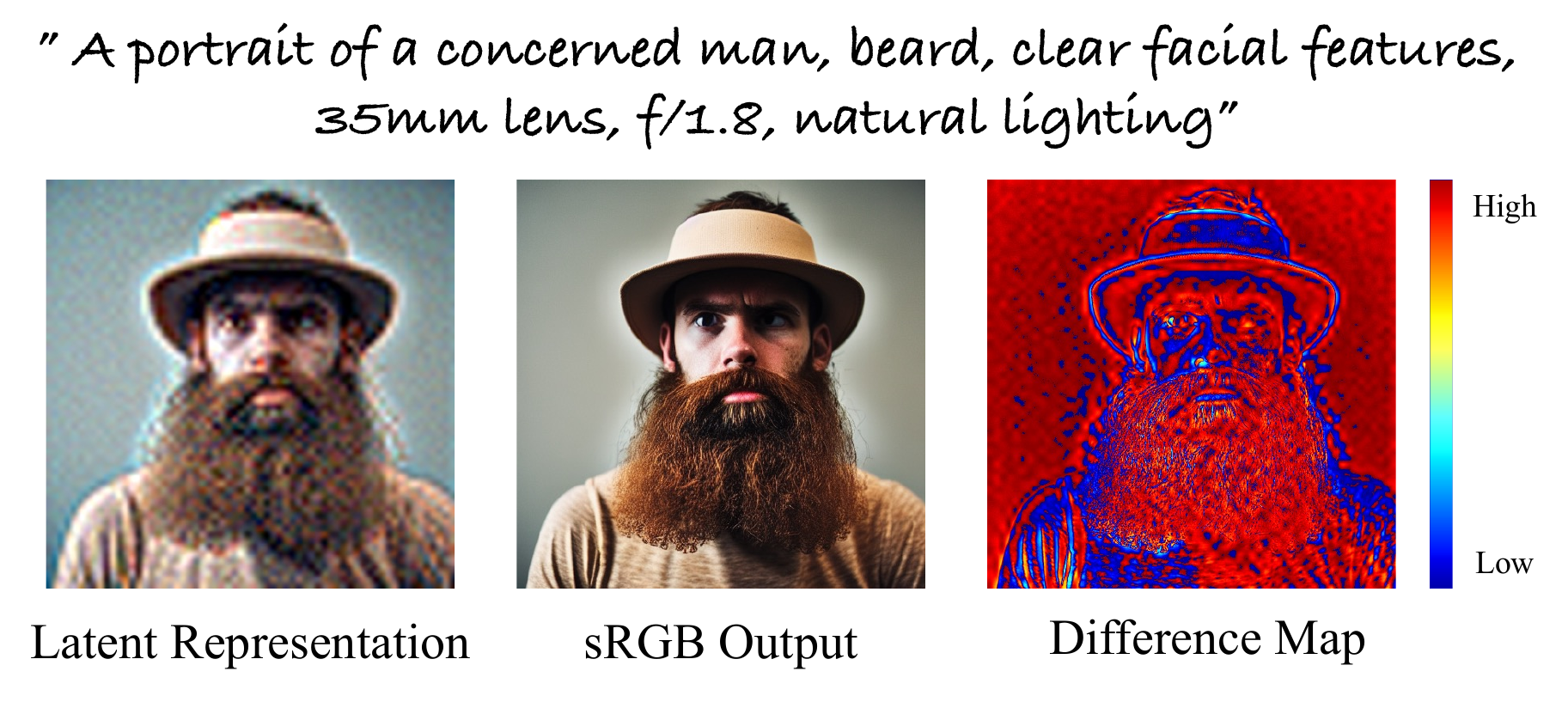}
\caption{A text-to-image example from Stable Diffusion. The difference map between the latent representation and sRGB output indicates that the latent representation generated by the UNet shows essential structural content while the decoder $\mathcal{D}$ introduces abundant details but scarcely modifies the global structure for the final sRGB output. The latent representation is converted to sRGB visualization by the linear transformation~\cite{Keturn2023}.}
\label{fig: priors}
\end{figure}

\subsection{Generative Priors in the Latent Diffusion Model}
% The diffusion model we adopt in this study is named Stable Diffusion~\cite{rombach2022high}.
We utilize the Stable Diffusion~\cite{rombach2022high} as our backbone. 
Stable Diffusion is a text-to-image latent diffusion model trained on the LAION-5B dataset~\cite{schuhmann2022laion} which is approximately 3M times bigger than the widely-used LLIE dataset SID-Sony~\cite{chen2018learning}.
It consists of an autoencoder and a UNet.
Delving into the autoencoder, the encoder $\mathcal{E}$ serves to encode an sRGB image into a latent representation $Z$ while the decoder $\mathcal{D}$ reconstructs the sRGB image from $Z$.
The UNet performs the denoising process for the noised latent representation $Z_{t}$, where $t$ denotes the timestep.

Recently, a diffusion model-based method~\cite{wang2023exploiting} is proposed to exploit generative priors of Stable Diffuiosn to benefit the image super-resolution task.
However, they do not delve into the presence of dedicated generative priors within different portions of Stable Diffusion.
With the further study of these priors within Stable Diffusion, we have two observations shown in Fig.~\ref{fig: priors}.
1) Given a text prompt, the UNet primarily serves to generate the latent representation which shows high-correlation structural information to the prompt and neglects to include intricate details;
2) During decoding the latent representation to the sRGB output, the decoder $\mathcal{D}$ introduces abundant details to improve the perceptual quality of the output but scarcely modifies the global structure.
These facts indicate that the UNet mainly contains the low-frequency generative prior while the decoder $\mathcal{D}$ chiefly has the high-frequency generative prior.

Based on the above observation that the UNet is adept at generating structural content, which is predominantly dedicated by low-frequency information, whereas the decoder demonstrates proficiency in the generation of high-frequency details as shown in Fig.~\ref{fig: priors}, we are inspired to extract different-frequency sub-bands from the input image by 2D discrete wavelet transforms (DWT) and separately feed them into the UNet and decoder $\mathcal{D}$ to exploit dedicated generative priors.
\subsection{Low-frequency Content Generation}
% \begin{figure}[!h]
% \centering
% \includegraphics[width=\linewidth]{../figs/DWT_visual_ICRA.pdf}
% \caption{The low-frequency sub-band, extracted using 2D discrete wavelet transforms (DWT), reveals clearer structural information compared to the low-light input.}
% \label{fig: DWT_visual}
% \end{figure}

Employing a series of DWT not only extracts the low-frequency sub-band (LL) but also allows us to efficiently align the input information with the latent size of the UNet.
Compared to directly resizing the input RAW image, feeding the low-frequency sub-band (LL) to the UNet enables the model to concentrate on generating high-correlation structural content while reducing the negative impact of the noise, as shown in Fig.~\ref{fig: framework} (b).

By initializing a latent representation $Z_{T} \sim \mathcal{N}(0, I)$, we aim at steering the denoising process in the UNet by inserting a taming module at each layer.
After denoising $T$ timesteps, we obtain a clean latent representation $Z_{0}$ showing a high correlation to the low-frequency sub-band (LL).

\paragraph{Taming Module} For the taming modules of the UNet, we employ a similar time-aware guidance strategy as discussed in~\cite{wang2023exploiting}.
However, instead of deriving the structural information by encoding the input image with the encoder $\mathcal{E}$ as in~\cite{wang2023exploiting}, we directly utilize the low-frequency sub-band (LL) to provide our structural information.
This liberates us from the substantial computational cost of encoding.
% This liberates us from the substantial computational cost of encoding, as shown in Table~\ref{tab: computational cost}.

% \begin{table}
%     \centering
%     \tabcolsep=0.03cm
%     \caption{A computational comparison between the encoder $\mathcal{E}$ adopted in~\cite{wang2023exploiting} and our adopted DWT.}\label{tab: computational cost}
%     \setlength\arrayrulewidth{1.0pt}
%     \resizebox{\linewidth}{!}{
%     \begin{tabular}{cccccc}
%         \toprule
%         \multicolumn{1}{p{1cm}}{}
%         &\multicolumn{1}{p{2cm}}{\centering Input size}
%         &\multicolumn{1}{p{1cm}}{\centering Time (s)}
%         &\multicolumn{1}{p{1cm}}{\centering Params (M)}
%         &\multicolumn{1}{p{1cm}}{\centering MACs (M)}
%         &\multicolumn{1}{p{2cm}}{\centering Finetune \\ (for RAW)} \\
%         \cmidrule(lr){1-6}
%         Encoder $\mathcal{E}$  & $512 \times 512$ & 2.09 & 34.16 & 566,371 & \ding{51}\\
%         \cmidrule(lr){1-6}
%         DWT & $512 \times 512$ & 0.02 & 0 & 2.625 & \ding{56} \\
%         \bottomrule
%     \end{tabular}}
% \end{table}

We insert a taming module at each layer in the UNet.
The taming module consists of a set of convolutional layers and an SFT layer for mapping the low-frequency sub-band (LL) to a pair of affine transformation parameters:
\begin{equation}
\alpha_{l}, \beta_{l} = \text{SFT}_{l}(\text{Conv}_{l}(\text{LL, t})),
\end{equation}
where $l$ denotes the layer number in the UNet and t denotes the timestep.
Then, the feature $\textbf{\text{F}}^{\text{UNet}}_{l}$ at the layer $l$ in the UNet is modulated by
\begin{equation}
\hat{\textbf{\text{F}}}^{\text{UNet}}_{l} = (1 + \alpha_{l}) \odot \textbf{\text{F}}^{\text{UNet}}_{l} + \beta_{l}.
\end{equation}

After taming the UNet to generate the latent representation $Z_{0}$, an sRGB can be directly obtained by feeding it to the decoder $\mathcal{D}$.
As shown in Fig.~\ref{fig: ablation_comparison}, such sRGB image presents well-exposed and high-correlation structural content but inaccurate color and over-generated details.
This inspires us to tame the decoder $\mathcal{D}$ during upsampling to guarantee exact illumination and precise details in the final sRGB image.

\subsection{High-frequency Detail Maintenance}
This high-frequency ``taming" aims to leverage the noise-free and fine-grained generative prior in the decoder to produce final sRGB images with enhanced details and devoided noise.

Similar to the taming process in the UNet, we insert a taming module at each upsampling layer in the decoder $\mathcal{D}$.
For taming the decoder $\mathcal{D}$, not only the high-frequency subbands are needed but also the low-frequency.
There are two key reasons for that:
\begin{itemize}
\renewcommand{\labelitemi}{\textbullet}
\item To ensure the final sRGB image exhibits the correct illumination, the low-frequency sub-band (LL) is mapped to the scale parameter in the SFT layer. This strategy enables a linear adjustment of the intensity of decoder feature values during the modulation.
\item For steering the generation with precise details during upsampling, the high-frequency sub-bands ([LH, HL, HH]) are mapped to the shift parameter in the SFT layer for adding detailed information onto the adjusted decoder feature.
\end{itemize}

\subsection{Training Objectives}
For training the taming modules of the UNet, the ground-truth sRGB image $Y$ is first encoded into a latent representation $Z_{0}$ via the encoder $\mathcal{E}$.
Then, the latent representation is corrupted with noise using the standard forward diffusion schedule~\cite{ho2020denoising}: $Z_{0}\overset{\epsilon}{\rightarrow}Z_{t}$, where the noise $\epsilon \sim \mathcal{N}(0, I)$ and the timestep $t\in \left [1, T \right ]$.
Since the UNet serves to predict the noise from the noised representation, the training objective is
\begin{equation}
\mathcal{L}_{\text{UNet}} = \mathbb{E}_{Z_{0}, \epsilon \sim \mathcal{N}(0, I), t}\left [ \left \| \epsilon - \epsilon_{\theta}(Z_{t}, t, \tau^{\text{UNet}}_{\phi}(\text{LL})) \right \|^{2}_{2} \right ],
\end{equation}
where $\epsilon_{\theta}(\cdot)$ denotes the UNet and $\tau^{\text{UNet}}_{\phi}(\cdot)$ denotes taming modules.

For training the taming modules of the decoder $\mathcal{D}$, a clean representation $\hat{Z_{0}}$ is generated by our tamed UNet for each low-light input image and the training objective is
\begin{equation}
\mathcal{L}_{\mathcal{D}} = \mathbb{E}_{\hat{Z}_{0}}\left [ \left | \mathcal{D}(\hat{Z}_{0}, \tau^{\mathcal{D}}_{\phi}(\text{LL}, \text{H}^{\ast})) - Y \right | \right ],
\end{equation}
where $\text{H}^{\ast}$ denotes the concatenation of high-frequency subbands (LH, HL, HH) and $\tau^{\mathcal{D}}_{\phi}(\cdot)$ denotes taming modules.

Note that all parameters from the UNet and decoder $\mathcal{D}$ are \textbf{frozen} and only our proposed taming modules are \textbf{trainable}.

\begin{figure*}
\centering
\includegraphics[width=0.9\linewidth]{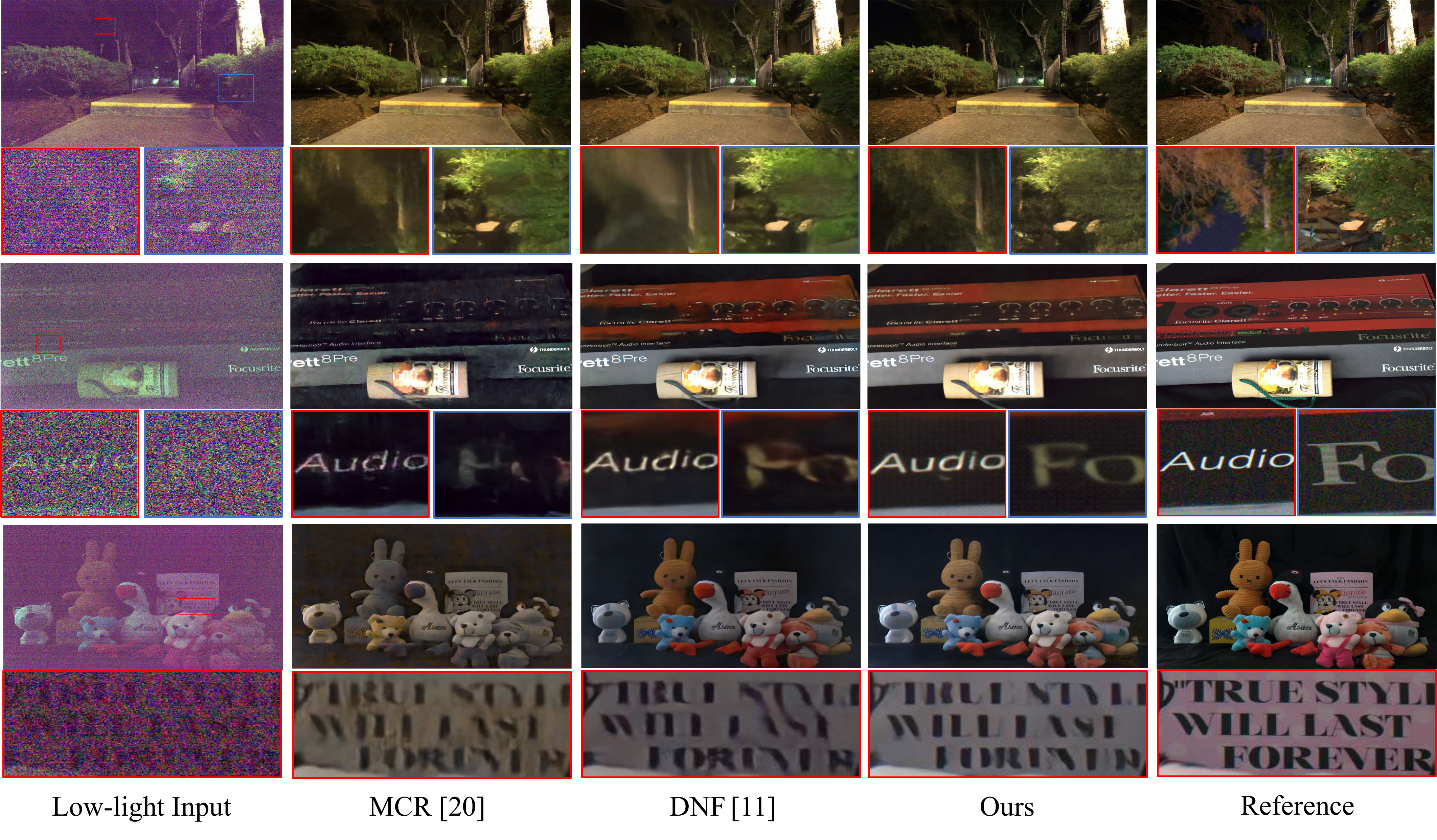}
\caption{The qualitative evaluations on SID-Sony~\cite{chen2018learning} and ELD-Sony~\cite{wei2021physics} datasets (Zoom-in for best view). The proposed method shows notable superiority in recovering structural information and enhancing details in extremely dark and noisy regions. (The illumination of the low-light input is increased for visualization.)}
\label{fig: visual_comparison_1}
\vspace{-5mm}
\end{figure*}

\section{Experiments}
\begin{table}
    \centering
    \tabcolsep=0.03cm
    \caption{The quantitative comparison in extremely challenging scenarios from different datasets. The proposed method achieves state-of-the-art performance according to perceptual metrics, including LPIPS~\cite{zhang2018perceptual} and NIMA~\cite{talebi2018nima}. It indicates the sRGB results obtained from the proposed method exhibit significantly clearer global structure and more natural details compared to those from previous works. The best score is denoted as \textbf{boldfaced}.($\dagger$ denotes the RAW-to-sRGB method and $\ast$ denotes the RAW-to-RAW method)}~\label{tab: score_comparison}
    \setlength\arrayrulewidth{1.0pt}
    \resizebox{\linewidth}{!}{
    \begin{tabular}{cccccc}
        \toprule
        \multicolumn{1}{p{2cm}}{\centering Dataset}
        &\multicolumn{1}{p{2.5cm}}{\centering Amplification Ratio}
        &\multicolumn{1}{p{2cm}}{\centering Methods}
        &\multicolumn{1}{p{1cm}}{\centering LPIPS $\downarrow$}
        &\multicolumn{1}{p{1cm}}{\centering NIMA $\uparrow$}
        &\multicolumn{1}{p{1cm}}{\centering SSIM $\uparrow$} \\
        \cmidrule(lr){1-6}
        \multirow{5}{*}{SID-Sony~\cite{chen2018learning}} & \multirow{5}{*}{$\times$300} & SID$^{\dagger}$~\cite{chen2018learning} & 0.3906 & 3.0481 & 0.8219 \\
     && ELD$^{\ast}$~\cite{wei2021physics} & 0.3876 & 3.0346 & 0.8271 \\
     && LED$^{\ast}$~\cite{jin2023lighting} & 0.4168 & 3.1831 & 0.8228 \\
     && MCR$^{\dagger}$~\cite{dong2022abandoning} & 0.5755 & 2.2367 & 0.7875 \\
     && DNF$^{\dagger}$~\cite{jincvpr23dnf} & 0.5145 & 1.9610 & \textbf{0.8494} \\
     && Ours$^{\dagger}$ & \textbf{0.2749} & \textbf{3.5583} & 0.8113 \\
        \cmidrule(lr){1-6}
        \multirow{5}{*}{ELD-Sony~\cite{wei2021physics}} & \multirow{5}{*}{$\times$200} & SID$^{\dagger}$~\cite{chen2018learning} & 0.2618 & 2.9780 & 0.7990 \\
     && ELD$^{\ast}$~\cite{wei2021physics} & 0.2585 & 3.1798 & 0.8063 \\
     && LED$^{\ast}$~\cite{jin2023lighting} & 0.2058 & 3.1718 & 0.8769 \\
     && MCR$^{\dagger}$~\cite{dong2022abandoning} & 0.3120 & 2.2043 & 0.7798 \\
     && DNF$^{\dagger}$~\cite{jincvpr23dnf} & 0.2255 & 2.0921 & 0.8683 \\
     && Ours$^{\dagger}$ & \textbf{0.2035} & \textbf{3.4397} & \textbf{0.8906} \\
        \cmidrule(lr){1-6}
        \multirow{2}{*}{LRD~\cite{zhang2023towards}} & \multirow{2}{*}{-3EV} & LRD$^{\ast}$~\cite{zhang2023towards} & 0.2198 & 3.1985 & 0.8784 \\
     && Ours$^{\dagger}$ & \textbf{0.1858} & \textbf{3.6180}& \textbf{0.8889} \\
        \bottomrule
    \end{tabular}}
    \vspace{-5mm}
\end{table}

% \begin{figure}[!th]
% \centering
% \includegraphics[width=\linewidth]{../figs/Visual_comparison_2_tmp.pdf}
% \caption{The qualitative evaluations on LRD~\cite{zhang2023towards} dataset (Zoom-in for best view).}
% \label{fig: visual_comparison_2}
% \end{figure}

\subsection{Implementation Details}
The proposed method is evaluated on three widely-used datasets for the LLIE task, including SID-Sony~\cite{chen2018learning}, ELD-Sony~\cite{wei2021physics}, and LRD~\cite{zhang2023towards}.
The train and test subsets follow the same strategy as previous works.
Given the ground-truth RAW images from these datasets, we obtain their corresponding sRGB images through the open-source ISP pipeline, rawpy\footnote{https://pypi.org/project/rawpy/}.
The same ISP pipeline is also adopted for the outputs of our RAW-to-RAW baselines.

During training, the taming modules of the UNet are trained for 100 epochs while the taming modules of the decoder are trained for 30 epochs.
By initially taming the UNet, it gains the capability to generate the clean latent representation $Z_{0}$ for each low-light training image.
This generated latent representation is subsequently utilized to train the decoder taming modules.
The training process is performed on 512 $\times$ 512 resolution with 4 NVIDIA GeForce RTX 3090 GPUs.
During inference, the UNet follows the DDIM sampling strategy~\cite{song2020denoising} with 200 timesteps.
\subsection{Baselines}
To demonstrate the superiority of the proposed~\method~on enhancing ISP for low-light images, we first conduct a comparison with strong RAW-to-sRGB LLIE methods.
To further highlight our denoising performance, we also compare the proposed method with RAW-to-RAW methods.
The state-of-the-art methods compared with the proposed \method~are: 1) RAW-to-sRGB methods including SID~\cite{chen2018learning}, MCR~\cite{dong2022abandoning} and DNF~\cite{jincvpr23dnf}; 2) RAW-to-RAW methods including ELD~\cite{wei2021physics}, LED~\cite{jin2023lighting} and LRD~\cite{zhang2023towards}.

\begin{figure*}[!h]
\centering
\includegraphics[width=0.9\linewidth]{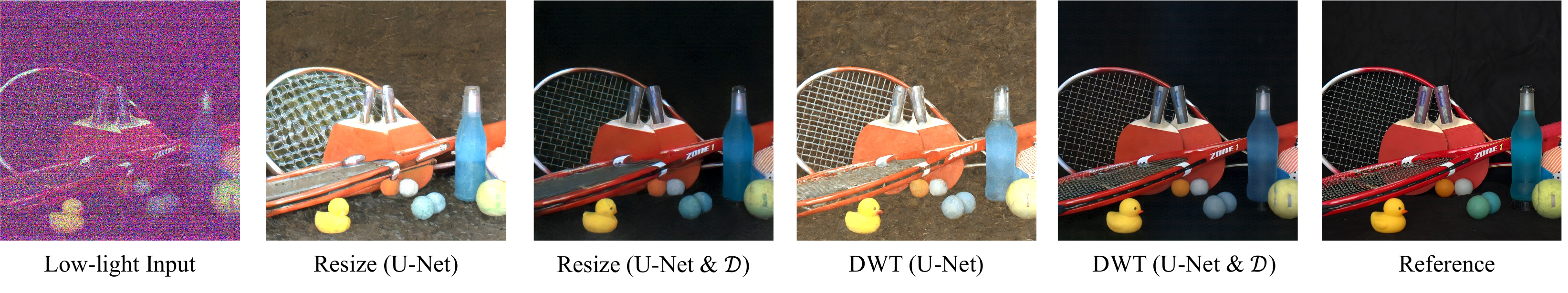}
\caption{The visualization of the ablation study. Different taming strategies (resize and DWT) and taming targets (UNet and decoder $\mathcal{D}$) are combined in this study. (The illumination of the low-light input is increased for visualization.)} % Taming the UNet by resizing the input information tends to result in an ambiguous global structure. In contrast, the taming result using 2D discrete wavelet transforms (DWT) exhibits distinct structural content. Moreover, taming the decoder $\mathcal{D}$ results in improved color distribution and the enhancement of details.
\label{fig: ablation_comparison}
\vspace{-3mm}
\end{figure*}

\subsection{Quantitative Evaluation}
In this work, we mainly focus on the perceptual quality of sRGB results which are enhanced from extremely dark and noisy RAW images.
Besides the metric SSIM, we also adopt LPIPS~\cite{zhang2018perceptual} and NIMA~\cite{talebi2018nima}for our quantitative evaluations.
LPIPS~\cite{zhang2018perceptual} and NIMA~\cite{talebi2018nima} are two widely-used reference-based and non-reference-based neural perceptual metrics.

According to Table ~\ref{tab: score_comparison}, the proposed method outperforms other methods in terms of neural perceptual metrics, LPIPS~\cite{zhang2018perceptual} and NIMA~\cite{talebi2018nima}.
To be specific, the proposed method achieves 0.1127 and 0.0023 improvements in LPIPS~\cite{zhang2018perceptual} than the previous sate-of-the-art methods on SID-Sony~\cite{chen2018learning} and ELD-Sony~\cite{wei2021physics} subsets, respectively.
Meanwhile, there are 0.3752 and 0.2599 improvements in NIMA~\cite{talebi2018nima} on two subsets, respectively.
It indicates the sRGB results obtained from the proposed method exhibit significantly more natural details compared to those from previous works.
The structural quality assessment based on SSIM and LPIPS~\cite{zhang2018perceptual} indicates the proposed method has substantial advantages in enhancing low-frequency information, especially structural content.
Note that the proposed method not only achieves state-of-the-art ISP performance compared to RAW-to-sRGB methods but also presents remarkably better denoising performance than RAW-to-RAW methods, which is a significant challenge to previous RAW-to-sRGB methods.

\subsection{Qualitative Evaluation}

Fig.~\ref{fig: visual_comparison_1} shows the visual comparison with other state-of-the-art methods on the SID-Sony~\cite{chen2018learning}, ELD-Sony~\cite{wei2021physics} datasets.
Images captured in extremely low-light conditions require high amplification ratios ($\times$300 or -3 EV).
These amplified images are plagued by acute noise and color distortion, making most details challenging to discern with the human eye.
Compared with previous works, the proposed method shows notable superiority in recovering structural information and enhancing details in noisy regions.
In the $\text{1}^{\text{st}}$ row of Fig.~\ref{fig: visual_comparison_1}, the results produced by previous works tend to be over-smooth due to the excessive denoising while the proposed method present sharp and clean leaves without color bias.
In the $\text{3}^{\text{rd}}$ row of Fig.~\ref{fig: visual_comparison_1}, previous works struggle to produce meaningful text information when portions of it are obscured by intense noise in the low-light image.
The proposed method, which leverages the potent generative prior from pre-trained Stable Diffusion, exhibits a significantly improved reconstruction of the text.
 %

% Thanks to the noise-free generative prior in the decoder $\mathcal{D}$, the sRGB images produced by our proposed method consistently present clean content, irrespective of the degree of noise in low-light inputs.

\subsection{Ablation Study}
\begin{table}
    \centering
    \tabcolsep=0.03cm
    \caption{The ablation study of the proposed method on ELD-Sony dataset~\cite{wei2021physics}. Different taming strategies (resize and DWT) and taming targets (UNet and decoder $\mathcal{D}$) are combined in this study. The best score is denoted as \textbf{boldfaced}.}\label{tab: ablation}
    \setlength\arrayrulewidth{1.0pt}
    \resizebox{\linewidth}{!}{
    \begin{tabular}{ccccc}
        \toprule
        \multicolumn{1}{p{2cm}}{}
        &\multicolumn{1}{p{2cm}}{\centering UNet}
        &\multicolumn{1}{p{2cm}}{\centering Decoder $\mathcal{D}$}
        &\multicolumn{1}{p{1.5cm}}{SSIM $\uparrow$}
        &\multicolumn{1}{p{1.5cm}}{LPIPS $\downarrow$} \\
        \cmidrule(lr){1-5}
        \multirow{2}{*}{Resize}  & \ding{51} & - & 0.7544 & 0.5423 \\
        \cmidrule(lr){3-5}
                                    & \ding{51} & \ding{51} & 0.8233 & 0.2834 \\
        \cmidrule(lr){1-5}
        \multirow{2}{*}{DWT} & \ding{51} & - & 0.7652 & 0.4904 \\
        \cmidrule(lr){3-5}
                             & \ding{51} & \ding{51} & \textbf{0.8906} & \textbf{0.2035} \\
        \bottomrule
    \end{tabular}}
    \vspace{-2mm}
\end{table}

\paragraph{Effectiveness of the DWT} 
% As we mentioned before, 2D discrete wavelet transforms (DWT) not only extract sub-bands for the taming but also downsample the input information to align with the latent size in the UNet.
Our 2D discrete wavelet transforms (DWT) not only extract sub-bands for the taming but also downsample the input information to align with the latent size in the UNet.
 %
% Instead of DWT, there are two other intuitive ways for downsampling: rescaling and CNN-based downsampling.
We also compare DWT with two other downsampling ways: rescaling and CNN-based downsampling.
 %
% In this study, we forego discussion on CNN-based downsampling due to its substantial computational expense.
 %
As shown in Fig.~\ref{fig: framework} (b), simply resizing the input (e.g., using Bicubic Interpolation) results in the loss of many details and causes the small-scale input to be overwhelmed by noise.
Taming the UNet with such input could result in an ambiguous global structure, potentially leading to low-quality sRGB images after decoding.
A qualitative comparison between resizing and DWT taming strategies is shown in Fig.~\ref{fig: ablation_comparison}.

\paragraph{About taming} 
% The taming modules of the UNet hold significant importance as they steer the generating process toward generating specified latent representations.
% The taming modules of the UNet are important as they steer the generating process toward generating specified latent representations.
%  %
% There is no need to conduct an ablation study on whether taming the UNet since the UNet without taming generates unexpected latent representation which is not related to the low-light input.
 
% The fact that motivates us to tame the decoder $\mathcal{D}$ is that although the low-frequency sub-band (LL) helps the proposed method to tame the UNet for generating a high-correlation latent representation, the sRGB image decoded from this representation by a decoder lacking taming modules, shows much over-generated and irrelevant details.
 
% Fig.~\ref{fig: ablation_comparison} and Table~\ref{tab: ablation} show that the decoder $\mathcal{D}$ when utilizing our DWT taming strategy, exhibits substantial improvements in color correction and high-frequency detail maintenance performance.

The taming modules of the UNet are crucial for guiding the generation process toward specific latent representations. An ablation study on taming is unnecessary since an untamed UNet produces irrelevant latent representations unrelated to low-light inputs.

We are motivated to tame the decoder $\mathcal{D}$ because, while the low-frequency sub-band (LL) aids in generating high-correlation latent representations, a decoder without taming modules results in over-generated and irrelevant details in the decoded sRGB image.

As shown in Fig.\ref{fig: ablation_comparison} and Table~\ref{tab: ablation}, using our DWT taming strategy significantly improves the decoder $\mathcal{D}$ in color correction and high-frequency detail maintenance.

\section{Conclusion}
To enhance neural ISP for the LLIE performance in extremely dark and noisy images, this work extricates itself from collecting more real-world data for training learning-based methods and designing more sophisticated network architecture.
Instead, we propose to tame a pre-trained diffusion model and explore its generative priors for achieving the LLIE task.
Our designed taming modules are fed with the low-light input information, and modulate intermediate features within the diffusion model, which allows us to steer the generating process.
With the observation of dedicated generative priors in different portions of the diffusion model, we propose to utilize 2D discrete wavelet transforms to divide the LLIE task into two parts: the low-frequency content generation and the high-frequency detail maintenance.
Benefiting from the abundant and dedicated generative priors in the diffusion model, extensive experiments demonstrate that taming the diffusion model allows us to achieve the optimized balance between image fidelity and perceptual quality, leading to state-of-the-art LLIE performance.

\section{Acknowledgement}
This project was supported by the National Key R\&D Program of China under grant number 2022ZD0161501.


%%%%%%%%%%%%%%%%%%%%%%%%%%%%%%%%%%%%%%%%%%%%%%%%%%%%%%%%%%%%%%%%%%%%%%%%%%%%%%%%

%%%%%%%%%%%%%%%%%%%%%%%%%%%%%%%%%%%%%%%%%%%%%%%%%%%%%%%%%%%%%%%%%%%%%%%%%%%%%%%%



% \section*{ACKNOWLEDGMENT}

% The preferred spelling of the word ÒacknowledgmentÓ in America is without an ÒeÓ after the ÒgÓ. Avoid the stilted expression, ÒOne of us (R. B. G.) thanks . . .Ó  Instead, try ÒR. B. G. thanksÓ. Put sponsor acknowledgments in the unnumbered footnote on the first page.

%%%%%%%%%%%%%%%%%%%%%%%%%%%%%%%%%%%%%%%%%%%%%%%%%%%%%%%%%%%%%%%%%%%%%%%%%%%%%%%%

{\small
\bibliographystyle{IEEEtran}
\bibliography{ref}
}

\end{document}